\documentclass{article} 
\usepackage{iclr2026_conference,times}

\usepackage{amsmath,amsfonts,bm}

\def\eqref#1{equation~\ref{#1}}

\def\1{\bm{1}}

\DeclareMathAlphabet{\mathsfit}{\encodingdefault}{\sfdefault}{m}{sl}
\SetMathAlphabet{\mathsfit}{bold}{\encodingdefault}{\sfdefault}{bx}{n}

\usepackage{hyperref}
\usepackage{url}
\usepackage{booktabs}
\usepackage{graphicx}
\usepackage{subcaption}
\usepackage{CJKutf8}

\title{DiT-Reward: Generative Representations for Text-to-Image Reward Modeling}

\author{
\normalfont
\makebox[0.96\textwidth][c]{%
\begin{tabular}{@{}c@{\hspace{1.6em}}c@{\hspace{1.6em}}c@{\hspace{1.6em}}c@{\hspace{1.6em}}c@{}}
\textbf{Yuanming Yang}$^{1,2}$ & \textbf{Guoqing Ma}$^{1}$ & \textbf{Bo Wang}$^{1,3}$ & \textbf{Yuan Zhang}$^{1}$ & \textbf{Wei Tang}$^{1}$
\end{tabular}%
}\\[2pt]
\makebox[0.96\textwidth][c]{%
\begin{tabular}{@{}c@{\hspace{2.2em}}c@{\hspace{2.2em}}c@{}}
\textbf{Chenyi Li}$^{1,4}$ & \textbf{Haoyang Huang}$^{1}$ & \textbf{Nan Duan}$^{1}$
\end{tabular}%
}\\[5pt]
\makebox[0.96\textwidth][c]{$^{1}$JD Explore Academy, JD.com, Beijing, China \quad $^{2}$Tsinghua University, Beijing, China}\\[2pt]
\makebox[0.96\textwidth][c]{$^{3}$Beijing Institute of Technology, Beijing, China \quad $^{4}$Peking University, Beijing, China}
}

\iclrfinalcopy 
\begin{document}

\maketitle
\lhead{\footnotesize DiT-Reward: Generative Representations for Text-to-Image Reward Modeling}

\begin{abstract}
Can representations learned for image generation also support the evaluation of generated images? We study text-to-image reward prediction as a downstream task of generative representation learning. To this end, we introduce \textbf{DiT-Reward}, which converts a pretrained text-to-image Diffusion Transformer into a reward model by processing near-clean image latents and aggregating text-conditioned image representations across transformer layers. Under the same training data mixture as HPSv3, DiT-Reward outperforms HPSv3 on all four evaluated preference benchmarks, reaching 85.6\% on HPDv2 and 77.6\% on HPDv3. When the generative backbone is frozen, a lightweight learned head can still extract meaningful preference predictions from its representations. Probing across depth further reveals that downstream reward performance is strongest in the middle-to-late layers and benefits from combining representations across different stages. We also observe consistent positive scaling with generative backbone capacity. Finally, when used to optimize Stable Diffusion 3.5 Large with Flow-GRPO, DiT-Reward outperforms HPSv3 along the matched training trajectory, with particularly clear gains in realism. Direct latent scoring also achieves a $1.65\times$ inference speedup over HPSv3 with comparable peak memory. These results show that pretrained generative DiTs provide transferable representations for reward modeling and policy optimization.

\end{abstract}

\section{Introduction}
\label{sec:intro}
\begingroup
\hyphenpenalty=10000
\exhyphenpenalty=10000
``What I cannot create, I do not understand.'' This maxim, attributed to Feynman, suggests a tight connection between generation and understanding. Modern text-to-image diffusion transformers appear to raise the same question from the opposite direction: they can generate increasingly realistic, complex, and semantically rich images, yet the representations learned by these models are rarely used to understand or evaluate the generated results, or to align generation with human preferences. This raises a direct question: can the internal representations learned by a generator be turned into a reward signal for evaluating the outputs of that same generator?

Reward models have become an important interface for evaluating and optimizing generative models, supporting automatic evaluation, reranking, reinforcement learning, and preference optimization~\citep{xu2023imagereward,kirstain2023pickapic,wu2023human,ma2025hpsv3,black2023training,fan2023dpok,wallace2024diffusiondpo,liu2025flowgrpo}. However, existing text-to-image reward models are typically built on separately pretrained discriminative vision--language encoders or general-purpose VLMs, such as CLIP, BLIP, and their successors~\citep{radford2021learning,li2022blip}. Modern DiT and MMDiT policies jointly process text and image tokens in a latent generative space, whereas reward evaluation typically requires the generated outputs to be decoded into pixels and then mapped into a separate vision--language representation space~\citep{peebles2023dit,esser2024sd3,stability2024sd35,blackforest2024flux}. Evaluation across these separate spaces not only introduces additional image decoding and visual encoding, but also separates the reward signal from the internal generative representations of the policy. This raises a direct question: can reward modeling operate in the native representation space of the generator?

To this end, we view reward prediction as a downstream task of generative representation learning and introduce \textbf{DiT-Reward}, a method that directly converts a pretrained text-to-image DiT into a reward model. Because a DiT is originally trained on noisy latents rather than completed images, we first encode an input image into the latent space used by the generator and apply a small near-clean perturbation to better match the DiT input distribution. We then extract text-conditioned image token representations from multiple MMDiT layers and map their pooled features to a scalar reward with a lightweight MLP. Under the same training data mixture as HPSv3, DiT-Reward outperforms HPSv3 on all four evaluated benchmarks and obtains the best results on HPDv2 and HPDv3. When the SD3.5-Large generative backbone is frozen and only a lightweight reward head is trained, the resulting text-conditioned image representations can already support human preference prediction without adapting the generative backbone. Increasing the generative backbone size further yields consistent gains across all four benchmarks.

Many technical reports on language model alignment adopt a common design in which the policy and reward model are derived from the same or closely related pretrained language models and then trained separately for generation and reward prediction~\citep{ouyang2022training,touvron2023llama,bai2023qwen,yang2023baichuan}. Although they do not share their resulting weights, this common initialization allows the reward model to inherit semantic representations, pretrained knowledge, and a modeling space closely related to those of the policy. Text-to-image reinforcement learning has not widely adopted this design and instead typically obtains rewards from an independently pretrained CLIP, BLIP, or VLM~\citep{black2023training,fan2023dpok,wallace2024diffusiondpo,liu2025flowgrpo}. DiT-Reward brings the same idea to text-to-image reinforcement learning: in our experiments, both DiT-Reward and the policy are derived from Stable Diffusion 3.5 Large and subsequently optimized for reward modeling and generation, respectively. With Flow-GRPO, external evaluations by GPT-5 and Gemini-3-Flash show that the policy trained with DiT-Reward outperforms its HPSv3-trained counterpart along the matched training trajectory; dimension-specific evaluations and qualitative comparisons further show clear improvements in realism and structural accuracy. Direct policy latent scoring is also $1.65\times$ faster than HPSv3 with comparable peak memory.

Our contributions are threefold. First, we propose DiT-Reward and show that a pretrained text-to-image DiT can be directly repurposed as an effective reward model, achieving leading or competitive results across major human preference benchmarks. Second, through probing with a frozen backbone, representation analysis across layers, and backbone scaling, we evaluate the transferability of pretrained generative representations to reward modeling and study how network depth and model capacity affect downstream reward prediction. Third, we validate DiT-Reward for text-to-image reinforcement learning when the reward model and policy share a common pretrained model origin, and demonstrate the efficiency advantage of direct reward evaluation in latent space.

\par
\endgroup

\section{Related Work}
\label{sec:related_work}
\paragraph{Human preference modeling and text-to-image evaluation.}
Human preference modeling is widely used for evaluating and aligning text-to-image models. Existing methods learn image-text preference scorers from CLIP, BLIP, or stronger VLM backbones for evaluation and reranking~\citep{xu2023imagereward,kirstain2023pickapic,wu2023hpsv2,ma2025hpsv3}. Recent work further decomposes preference into fine-grained dimensions or adopts generative VLMs to improve interpretability, scalability, and robustness~\citep{xu2024visionreward,he2024videoscore,fang2025rewarddance}. In contrast, DiT-Reward reuses the text-to-image generative model itself and tests whether its internal representations are sufficient for human preference prediction.

\paragraph{Feedback learning for diffusion and flow models.}
Another line of work optimizes diffusion or flow generators with preference and reward signals, including reward-gradient guidance, policy-gradient formulations, KL-regularized online RL, and diffusion variants of DPO~\citep{prabhudesai2024aligning,black2023training,fan2023dpok,wallace2024diffusiondpo}. Recent methods extend this direction to flow matching and rectified-flow models, such as Flow-GRPO and DiffusionNFT~\citep{liu2025flowgrpo,zheng2025diffusionnft}. These works mainly ask how to optimize a generator given a reward. Our focus is complementary: we study how the reward model itself should be constructed.

\paragraph{Generative backbones as transferable representation learners.}
A growing body of work shows that pretrained generative models are also transferable visual representation learners. Diffusion features have been used for semantic correspondence, segmentation, depth estimation, and related dense prediction tasks, with quality depending on layer, timestep, noise level, and conditioning~\citep{tang2023dift,luo2023diffusionhyperfeatures,xu2023openvocabulary,zhao2023unleashing,stracke2025cleandift}. Meanwhile, text-to-image backbones have scaled from LDMs and DiTs to large multimodal diffusion and flow transformers~\citep{rombach2022ldm,peebles2023dit,podell2023sdxl,esser2024sd3,blackforest2024flux,wu2025qwenimage,cao2025hunyuanimage}. DiT-Reward connects these threads by using the generative backbone as the representation source for reward modeling.

\section{Method}
\label{sec:method}
\subsection{Problem Formulation}

We consider the problem of learning a reward model for text-to-image generation. Given a text prompt $p$ and a pair of generated images $(x^+, x^-)$, where $x^+$ is preferred over $x^-$ according to human judgments, our goal is to learn a reward function $r_\theta(x, p)$ that assigns higher scores to preferred images. Formally, we seek to learn $r_\theta$ such that $r_\theta(x^+, p) > r_\theta(x^-, p)$ for preference pairs in the training data. The learned reward model can then be applied to image ranking or as a reward signal for downstream reinforcement learning optimization of text-to-image models.

\begin{figure}[!t]
    \centering
    \includegraphics[page=1,width=\linewidth]{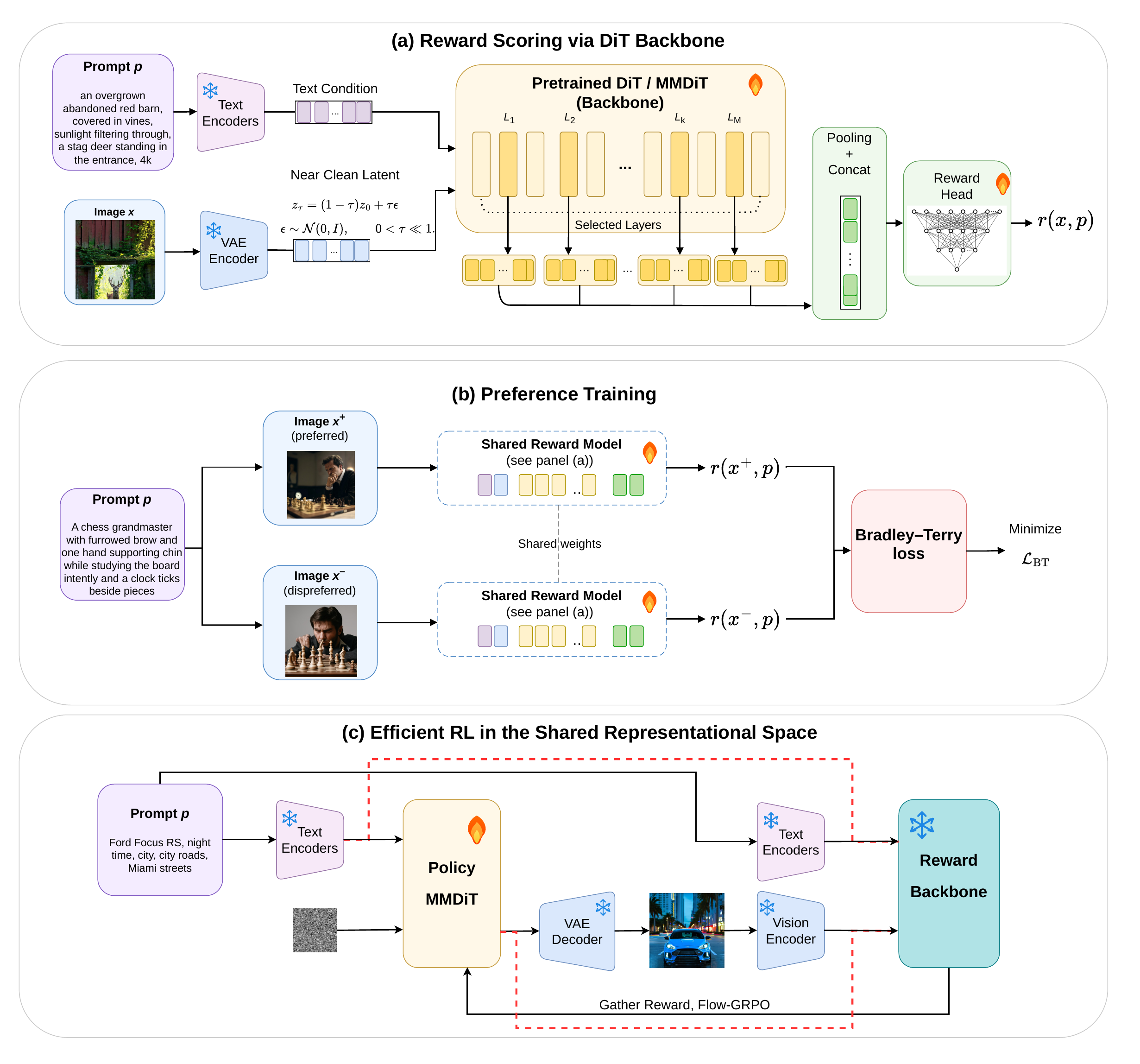}
    \caption{Model architecture, preference training, and reinforcement learning interface of DiT-Reward. \textbf{(a)} Given a prompt and image pair, the VAE maps the image into latent space, and a small amount of noise produces a nearly clean latent. A pretrained DiT or MMDiT jointly processes the text condition and image latent. Image token representations from selected layers are pooled, concatenated, and mapped to a scalar reward by a lightweight reward head. \textbf{(b)} For preferred and rejected images under the same prompt, DiT-Reward with shared parameters predicts two rewards and learns their ordering through the Bradley--Terry loss. \textbf{(c)} During reinforcement learning, a conventional pixel interface requires VAE decoding followed by visual encoding. When the policy and DiT-Reward share a latent space, the red path sends the generated latent directly to the reward model, avoiding additional decoding and encoding while reducing inference cost and peak memory.}
    \label{fig:method_overview}
\end{figure}

\subsection{Generator-Native Reward Modeling with DiT}

Most existing text-to-image reward models are built on discriminative vision-language encoders, such as CLIP-style or VLM backbones. DiT-Reward instead uses the pretrained text-to-image generator itself as the reward backbone. The motivation is that a DiT-based generator must internally model visual quality, prompt-image alignment, composition, and generation feasibility in order to synthesize images. Rather than introducing a separate vision-language encoder, we reuse these generator-native representations for reward prediction.

Given a prompt-image pair $(p,x)$, the prompt is encoded by the frozen text encoders of the text-to-image model, and the image is encoded into the VAE latent space. Since the DiT backbone is trained to process noisy latent states rather than perfectly clean image latents, directly feeding the clean latent would introduce an input-distribution mismatch. We therefore construct a near-clean latent by adding a small amount of Gaussian noise:
\begin{equation}
    z_\tau = (1-\tau) z_0 + \tau \epsilon, \quad \epsilon \sim \mathcal{N}(0, I), \quad 0 < \tau \ll 1,
\end{equation}
where $z_0$ denotes the VAE latent of $x$ and $\tau$ is a small constant. This perturbation places the image latent in a flow-time state that is more compatible with the pretrained DiT. The noise level must remain small: large noise would corrupt the image content and make the reward model score a substantially perturbed image rather than the original candidate.

The near-clean image latent and the text embeddings are then processed jointly by the pretrained MMDiT backbone. Let $\mathcal{S} = \{\ell_1, \ell_2, \ldots, \ell_K\}$ denote the selected DiT layers. For each layer $\ell \in \mathcal{S}$, we extract the image-token hidden states:
\begin{equation}
    H_\ell(x, p) \in \mathbb{R}^{N \times d},
\end{equation}
where $N$ is the number of image tokens (spatial positions in the latent space) and $d$ is the hidden dimension of the MMDiT.

We aggregate each layer's image-token representation by mean pooling:
\begin{equation}
    f_\ell(x, p) = \frac{1}{N} \sum_{i=1}^{N} H_\ell^{(i)}(x, p),
\end{equation}
and concatenate the pooled features across layers:
\begin{equation}
    f(x, p) = \mathrm{Concat}\left(f_{\ell_1}(x, p), f_{\ell_2}(x, p), \ldots, f_{\ell_K}(x, p)\right) \in \mathbb{R}^{K \cdot d}.
\end{equation}
This multi-layer aggregation exposes reward-relevant information from different depths of the generative model. The concatenated feature is then passed to a lightweight MLP reward head, which progressively projects the high-dimensional DiT representation into a scalar reward:
\begin{equation}
    r_\theta(x, p) = \mathrm{MLP}(f(x, p)) \in \mathbb{R}.
\end{equation}

\subsection{Preference Learning Objective}

We train DiT-Reward with the standard Bradley-Terry preference objective. Given a prompt $p$ and a preferred/rejected image pair $(x^+, x^-)$, the two images are scored by the same reward model. We then optimize
\begin{equation}
    \mathcal{L}(\theta) = -\log \sigma\left(r_\theta(x^+, p) - r_\theta(x^-, p)\right),
\end{equation}
where $\sigma(\cdot)$ is the sigmoid function. This loss increases the reward margin between human-preferred and rejected images. In our main setting, the text encoders and VAE remain frozen, while the DiT backbone and reward head can be fine-tuned for preference prediction; we also study a frozen-backbone variant in the ablation section to probe how much reward-relevant information already exists in the pretrained generator.

\subsection{Reinforcement Learning in the Latent Space}

As a general reward model, DiT-Reward can be used in the same way as conventional image reward models: a policy model first generates a complete image, and the reward model scores the resulting prompt-image pair. This makes DiT-Reward directly applicable to reranking and reward-based policy optimization.

Its generator-native design also enables a more efficient pathway when the policy model and reward model share the same latent space. Conventional VLM-based reward models require the policy latent to be decoded into pixels and then processed again by an external vision encoder. In contrast, DiT-Reward can score candidates from the shared latent representation after generation, avoiding unnecessary decoding and re-encoding steps. This latent-space reward evaluation reduces redundant computation and memory usage, which is especially useful for reinforcement learning with large diffusion or flow policies.

\section{Experiment}
\label{sec:experiment}
\subsection{Experimental Setup}
\label{subsec:exp_setup}

We instantiate DiT-Reward from Stable Diffusion 3.5 Large by reusing its MMDiT backbone as the reward feature extractor. This backbone has approximately 8B parameters and consists of 38 joint transformer layers that process VAE image-latent tokens together with text tokens formed from the concatenated embeddings of CLIP-L, OpenCLIP bigG, and T5-XXL. In our default implementation, image-token hidden states are extracted from layers 15, 23, 28, and 37, where the layer index starts from 0, and the pooled multi-layer representation is fed into a lightweight MLP reward head. Additional backbone configuration details are provided in Appendix~\ref{app:implementation_details}.

To focus the comparison on reward-model architecture rather than supervision, we adopt the public HPSv3 training recipe. The training mixture includes HPDv3 pairwise data, a filtered golden subset from HPDv2, sampled Pick-A-Pic data, sampled ImageReward data, and Midjourney user-choice data, without introducing additional data sources.

During training, images are encoded into the SD3.5 latent space and perturbed with Gaussian noise using $\tau = 0.005$. We train for 2--3 epochs with a warmup ratio of 0.05. The VAE and text encoders remain frozen. In the main setting, the DiT backbone and reward head are trainable, with learning rates $5\times 10^{-6}$ and $2\times 10^{-5}$, respectively. The reward head has $\sim$4M--16M parameters depending on layer count; Appendix~\ref{app:training_dynamics} reports training and benchmark trajectories.

\noindent\begin{minipage}{\linewidth}
\centering
\captionsetup{hypcap=false}
\captionof{table}{Main reward-model benchmark results. We report pairwise preference prediction accuracy (\%) on ImageReward, PickScore, HPDv2, and HPDv3. Best results are shown in bold and second-best results are underlined. For the HPSv3 PickScore entry marked by $^{*}$, we evaluate the official released HPSv3 checkpoint with the official PickScore evaluation protocol.}
\label{tab:rm_bench}
\begin{tabular}{lcccc}
\toprule
\textbf{Model} & \textbf{ImageReward} & \textbf{PickScore} & \textbf{HPDv2} & \textbf{HPDv3} \\
\midrule
CLIP ViT-H/14        & 57.1 & 60.8 & 65.1 & 48.6 \\
Aesthetic Score Predictor & 57.4 & 56.8 & 76.8 & 59.9 \\
ImageReward          & 65.1 & 61.1 & 74.0 & 58.6 \\
PickScore            & 61.6 & \textbf{70.5} & 79.8 & 65.6 \\
HPS                  & 61.2 & \underline{66.7} & 77.6 & 63.8 \\
HPSv2                & 65.7 & 63.8 & 83.3 & 65.3 \\
MPS                  & \textbf{67.5} & 63.1 & 83.5 & 64.3 \\
HPSv3                & 66.8 & 65.1$^{*}$ & \underline{85.4} & \underline{76.9} \\
DiT-Reward (Ours)    & \underline{67.0} & \underline{66.7} & \textbf{85.6} & \textbf{77.6} \\
\bottomrule
\end{tabular}
\end{minipage}\par

\subsection{Main Results on Preference Benchmarks}
\label{subsec:main_results}

\begingroup
\brokenpenalty=10000
Table~\ref{tab:rm_bench} compares DiT-Reward with representative reward models on pairwise preference prediction benchmarks. DiT-Reward achieves the best performance on HPDv3, reaching 77.6\%. We regard HPDv3 as the most important benchmark in this comparison because it contains images generated by more recent text-to-image models and is therefore more indicative of current reward-model performance. DiT-Reward also achieves the best result on HPDv2 with 85.6\%, ranks second on ImageReward, and ties for second on PickScore. Notably, DiT-Reward outperforms HPSv3 on all four benchmarks while using the same training data mixture.
\par
\endgroup

\subsection{Reinforcement Learning with DiT-Reward}
\label{subsec:rl_dit_reward}

We evaluate DiT-Reward as a reward signal for text-to-image reinforcement learning using Flow-GRPO~\citep{liu2025flowgrpo}. We use Stable Diffusion 3.5 Large as the policy and follow the official Flow-GRPO PickScore prompt configuration for both RL training and evaluation. Training images are generated at $512\times512$ resolution with 10 sampling steps and guidance scale 4.5, while evaluation uses 40 sampling steps. We use a group size of 16 and, through distributed gradient accumulation across 16 GPUs, an effective optimization batch size of 384 samples per update. The policy is optimized with LoRA, with $\beta=0.01$ controlling the strength of KL regularization. The preceding policy, sampling, and optimization settings are matched between the DiT-Reward and HPSv3 runs, while each reward model retains its native scoring pathway; in particular, DiT-Reward scores samples in latent space.

\begin{figure}[t]
\centering
\newcommand{\rlplaceholder}{%
    \fbox{%
    \begin{minipage}[c][0.15\linewidth][c]{0.16\linewidth}
    \centering
    \setlength{\tabcolsep}{0pt}
    \renewcommand{\arraystretch}{0}
    \newcommand{\bsq}{\rule{0.23\linewidth}{0.23\linewidth}}
    \newcommand{\wsq}{\phantom{\rule{0.23\linewidth}{0.23\linewidth}}}
    \begin{tabular}{@{}cccc@{}}
    \bsq & \wsq & \bsq & \wsq \\
    \wsq & \bsq & \wsq & \bsq \\
    \bsq & \wsq & \bsq & \wsq \\
    \wsq & \bsq & \wsq & \bsq
    \end{tabular}
    \end{minipage}}}
\newcommand{\rlimage}[1]{%
    \begin{minipage}[c][0.216\linewidth][c]{0.216\linewidth}
    \centering
    \includegraphics[width=0.96\linewidth]{#1}
    \end{minipage}}
\newcommand{\rlprompt}[1]{%
    \begin{minipage}[c][0.16\linewidth][c]{0.24\linewidth}
    \raggedright\emph{#1}
\end{minipage}}
\small
\setlength{\tabcolsep}{0pt}
\begin{tabular}{@{}l@{\hspace{10pt}}ccc@{}}
\toprule
\textbf{Prompt} & \textbf{Base policy} & \textbf{RL with HPSv3} & \textbf{RL with DiT-Reward} \\
\midrule
\rlprompt{Beautiful blonde girl laughing, profile view.} &
\rlimage{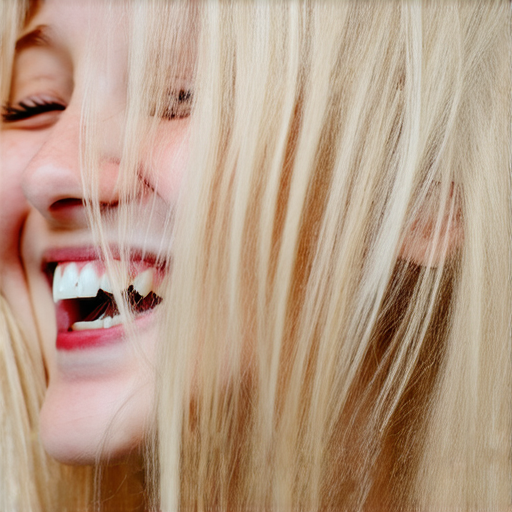} &
\rlimage{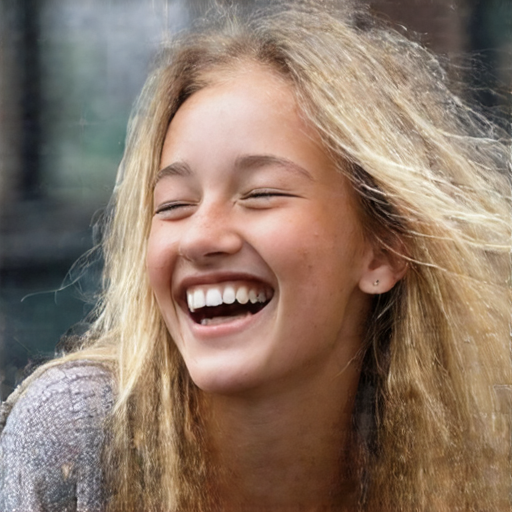} &
\rlimage{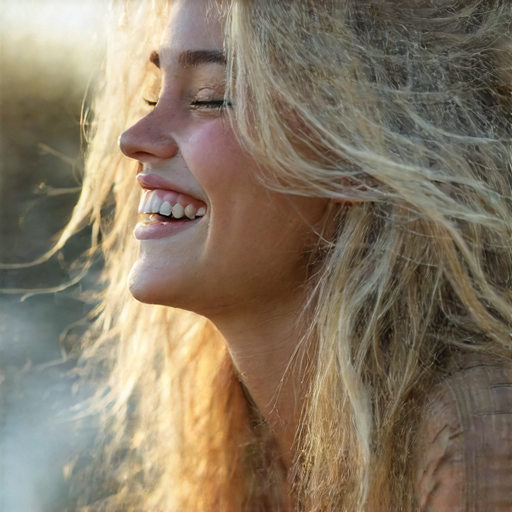} \\
\rlprompt{A happy Nepalese girl in a village.} &
\rlimage{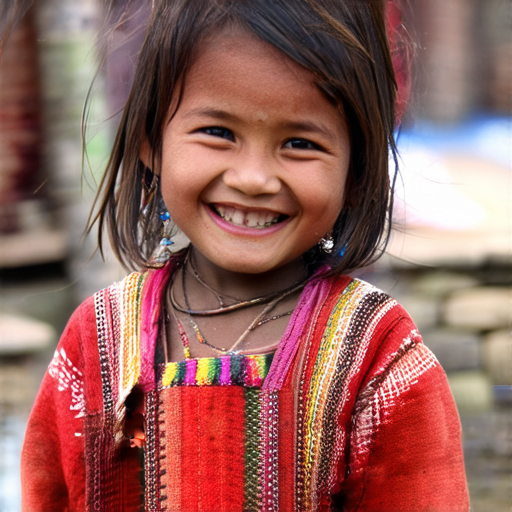} &
\rlimage{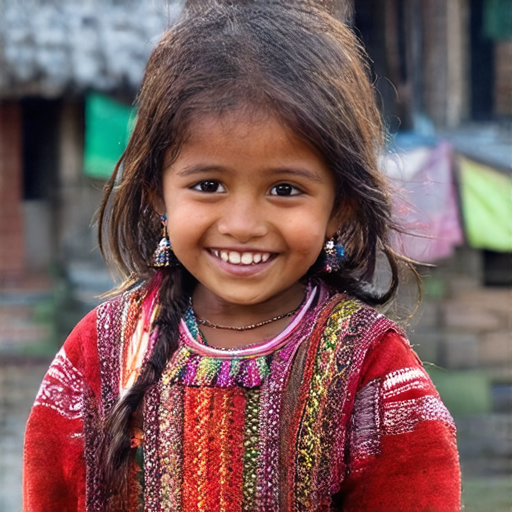} &
\rlimage{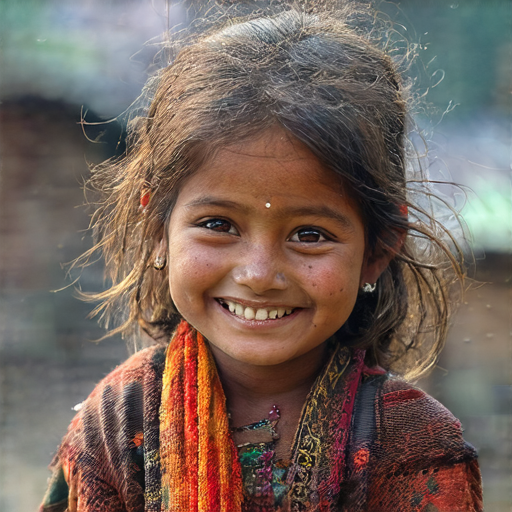} \\
\rlprompt{Purple dragon flying over a castle on a hill.} &
\rlimage{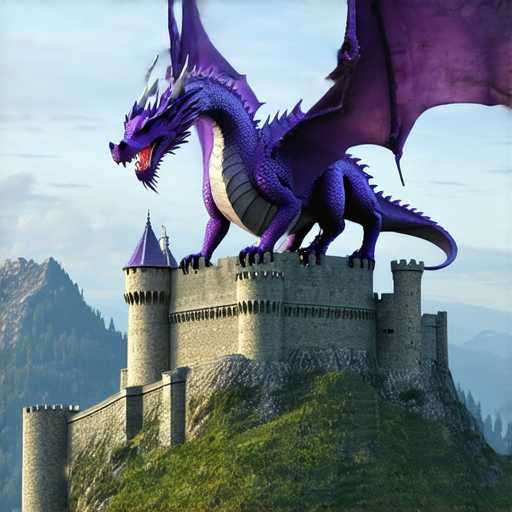} &
\rlimage{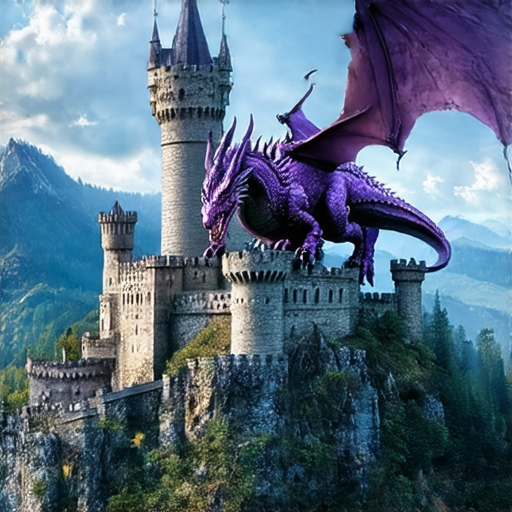} &
\rlimage{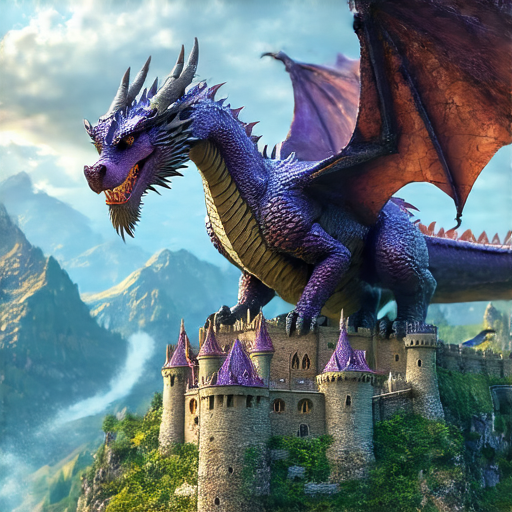} \\
\rlprompt{A Ferrari car that is made out of wood.} &
\rlimage{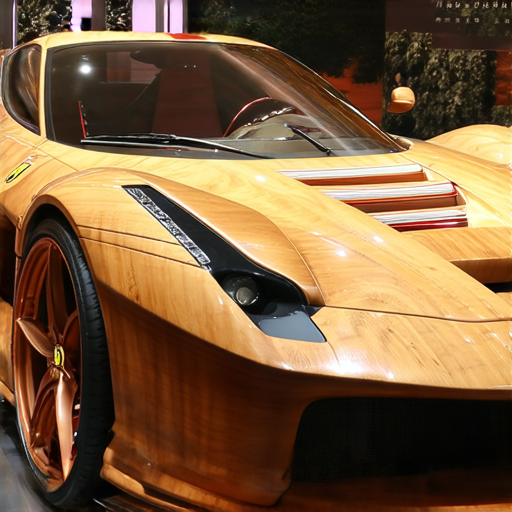} &
\rlimage{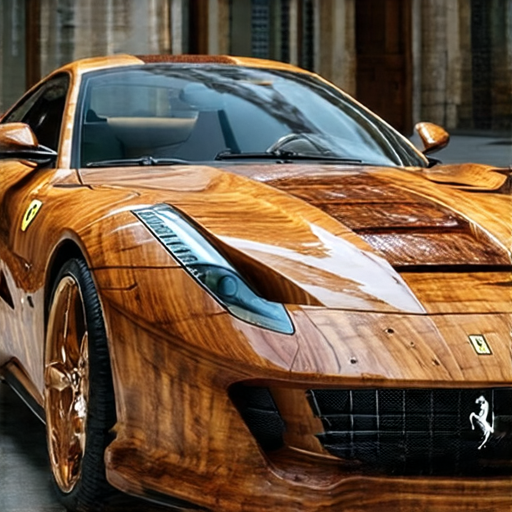} &
\rlimage{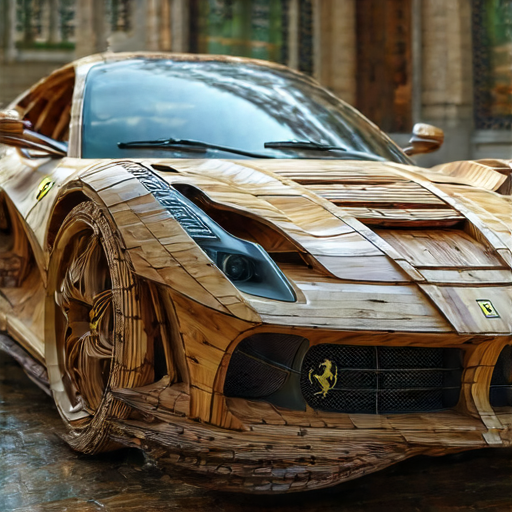} \\
\rlprompt{A titan ravaging the raging seas.} &
\rlimage{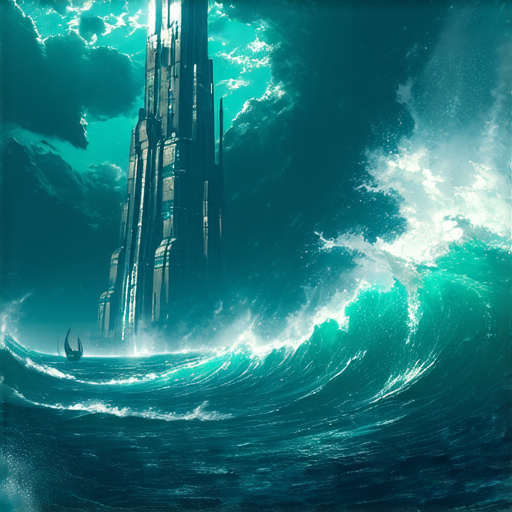} &
\rlimage{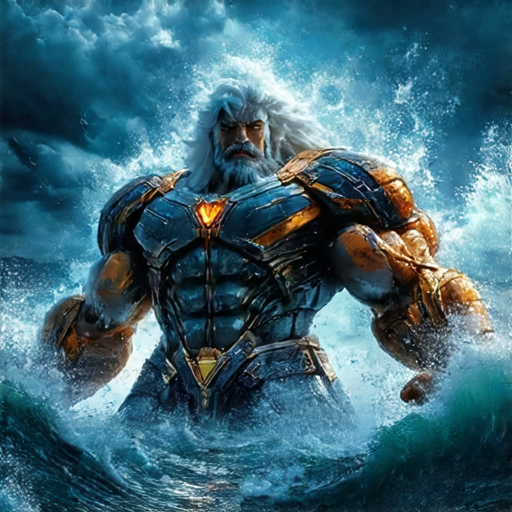} &
\rlimage{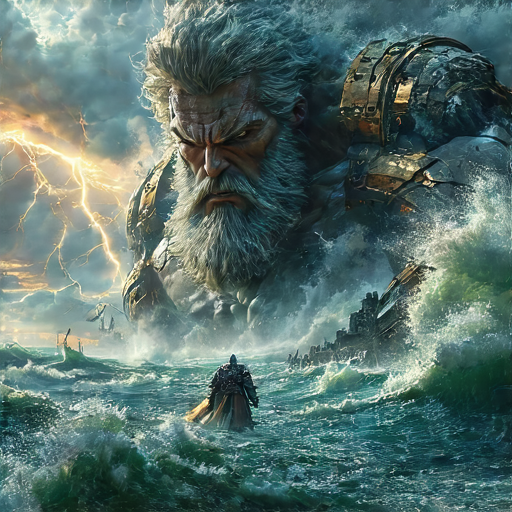} \\
\rlprompt{Abraham Lincoln gives an outdoor speech to large crowd.} &
\rlimage{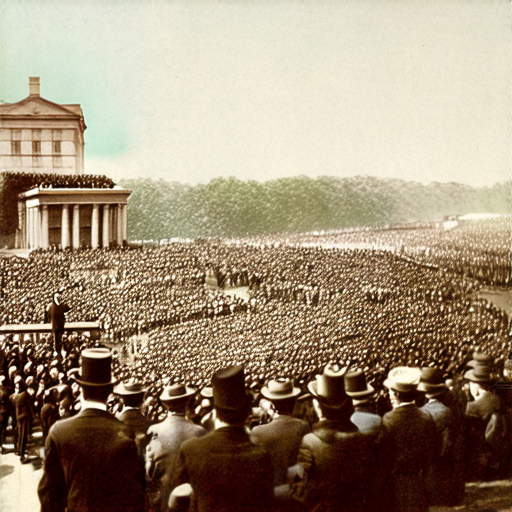} &
\rlimage{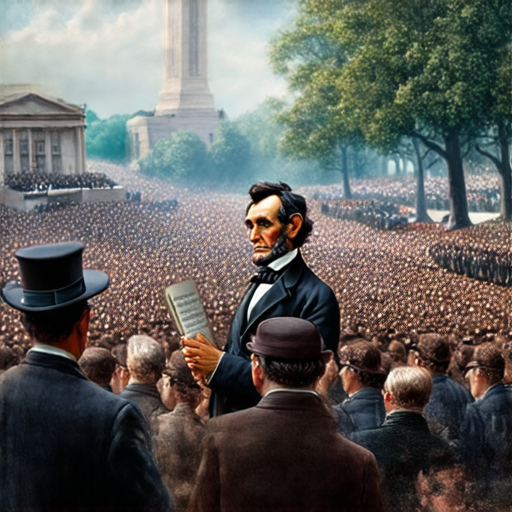} &
\rlimage{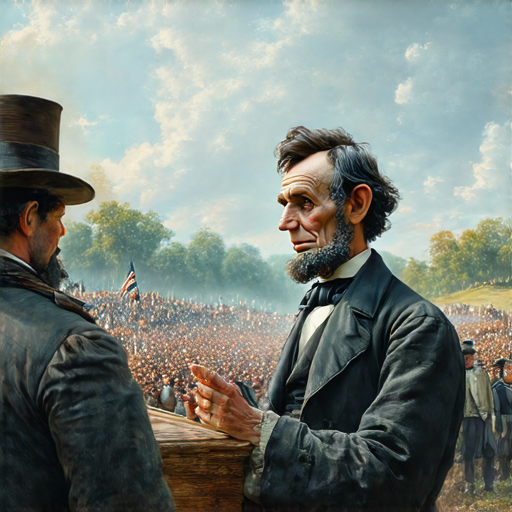} \\
\bottomrule
\end{tabular}
\caption{Qualitative comparison using the best-performing checkpoint from each training run. All examples from a given policy are sampled from the same selected checkpoint, and each row uses the same prompt to compare the base policy with policies optimized using HPSv3 or DiT-Reward.}
\label{fig:rl_qualitative}
\end{figure}

For external evaluation, we use the 2,048 PickScore prompts from the official Flow-GRPO protocol. GPT-5 and Gemini-3-Flash evaluate each generated image independently at 60-step intervals along four dimensions: prompt alignment, visual quality, realism, and detail richness. The complete prompts and evaluation protocol are provided in Appendix~\ref{app:rl_dimensions}. For qualitative analysis, we select the best-performing checkpoint from each training run and use the same selected checkpoint for all examples from that policy in Figure~\ref{fig:rl_qualitative}.

\begin{center}
    \centering
    \includegraphics[width=\linewidth]{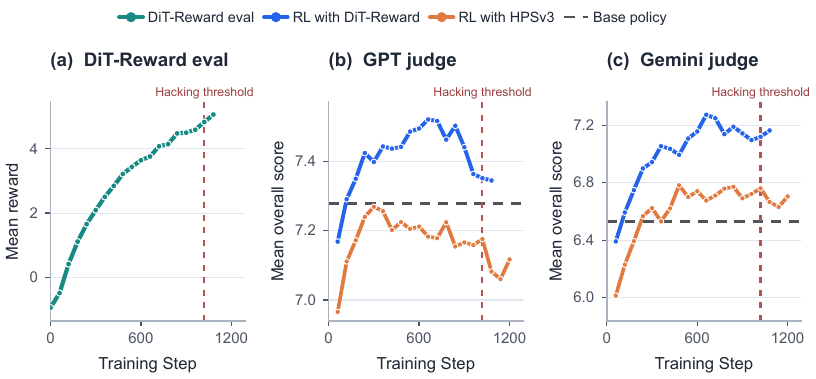}
    \captionsetup{hypcap=false}
    \captionof{figure}{Internal and external evaluation during Flow-GRPO training. Panel (a) shows mean DiT-Reward on evaluation samples for the DiT-Reward-trained policy. Panels (b) and (c) compare overall scores from GPT-5 and Gemini-3-Flash for policies trained with DiT-Reward (blue) and HPSv3 (orange). Gray dashed lines denote base-policy scores of 7.29 and 6.53, respectively. Red dashed lines mark step 1020, used as a common reference because visible reward-hacking artifacts emerge at a similar stage in both RL runs.}
    \label{fig:rl_vlm_overall}
\end{center}

As shown in Figure~\ref{fig:rl_vlm_overall}, Flow-GRPO steadily increases the DiT-Reward evaluation score from $-0.94$ to $5.06$, confirming that the policy can optimize the learned reward. Across the 18 matched evaluation checkpoints from steps 60 to 1080, replacing HPSv3 with DiT-Reward increases the average overall score from 7.18 to 7.41 under GPT-5 and from 6.61 to 7.02 under Gemini-3-Flash. DiT-Reward achieves a higher overall score at every matched checkpoint under both judges. The gains are concentrated in visual quality, realism, and detail richness, while HPSv3 retains a modest advantage in prompt alignment, particularly during later training. For the DiT-Reward run, the internal reward continues to rise after the external scores peak. Qualitative inspection indicates that visible reward hacking emerges at a similar stage in both reward-model runs, so step 1020 is shown as a common reference threshold. Appendix~\ref{app:rl_dimensions} reports complete trajectories for all four dimensions.

Figure~\ref{fig:rl_qualitative} compares matched-prompt generations from the base policy and policies optimized with HPSv3 and DiT-Reward. Before pronounced reward overoptimization, the policy optimized with DiT-Reward produces richer local details and more realistic portraits, animals, and objects. It also exhibits less drift toward excessive brightness, saturation, and contrast than the policy optimized with HPSv3, avoiding some unnatural colors and distorted shapes.

Both reward models eventually exhibit reward hacking: faces and objects become frontal and rigid, hallucinated details accumulate, and outputs converge toward reward-favored styles at the expense of prompt fidelity. Their failure modes nevertheless differ. HPSv3 tends to amplify saturation and contrast, sometimes producing implausible colors and geometry, whereas DiT-Reward better preserves global structure but may yield muted colors, spurious faces, and weaker prompt alignment. These differences indicate that reward models trained on the same preference data can induce distinct optimization biases because of their pretraining objectives and backbone architectures. A plausible explanation is that the diffusion backbone provides stronger structural priors, while the VLM-based reward retains an advantage in text-instruction following. One possible direction for future work is to incorporate text-token representations from the joint transformer and explicitly model text--image interactions to improve the prompt-alignment sensitivity of DiT-Reward.

\subsection{Efficient Reward Inference}
\label{subsec:reward_efficiency}

A key advantage of DiT-Reward is that it can directly evaluate generated latents when the reward model and policy share the same latent space. This pathway avoids VAE decoding and pixel-space re-encoding by an external vision backbone. We compare its inference efficiency with HPSv3 and the pixel-interface DiT-Reward pathway in Table~\ref{tab:reward_efficiency}.

\begin{center}
\captionsetup{hypcap=false}
\captionof{table}{Reward inference efficiency on $512\times512$ images measured on one NVIDIA RTX PRO 6000 Blackwell GPU with batch size 1. Each mode uses one warmup iteration excluded from timing. The reward interface indicates whether the model receives decoded pixels or policy latents. Lower values are better.}
\label{tab:reward_efficiency}
\small
\setlength{\tabcolsep}{5pt}
\begin{tabular}{lccc}
\toprule
\textbf{Reward Model} & \textbf{Reward Interface} & \textbf{Per-Image Latency} & \textbf{Peak Memory} \\
\midrule
DiT-Reward & Pixel  & 154 ms & 27.2 GB \\
HPSv3      & Pixel  & 89 ms  & \textbf{16.0 GB} \\
DiT-Reward & Latent & \textbf{54 ms} & 16.1 GB \\
\bottomrule
\end{tabular}
\end{center}

Although the complete DiT-Reward inference stack contains an 8B DiT backbone and approximately 5B additional parameters in its frozen text encoders, the latent-interface pathway achieves the lowest inference latency: 54 ms per image compared with 89 ms for HPSv3, corresponding to a $1.65\times$ speedup. Its peak memory consumption is 16.1 GB, nearly identical to the 16.0 GB required by HPSv3. DiT-Reward with the pixel interface instead requires 154 ms per image and 27.2 GB of memory. Together with the improved reward-model and downstream RL performance reported above, these results indicate that aligning the policy and reward model in a shared latent representation space can improve performance without sacrificing inference efficiency. In particular, the aligned latent interface enables a large generative reward model to achieve faster inference with a memory footprint comparable to HPSv3, highlighting representation-space alignment as both a modeling and systems advantage for text-to-image reinforcement learning.

\section{Ablation and Interpretability}
\label{sec:ablation}
We organize our analysis around two objectives. First, we examine whether representations from a pretrained generative DiT can support reward modeling with a frozen backbone and identify which layers are most informative for this task. Second, we investigate whether reward modeling performance exhibits scaling behavior as the parameter count of the generative backbone increases.

\begin{figure}[!t]
    \centering
    \includegraphics[width=0.80\linewidth]{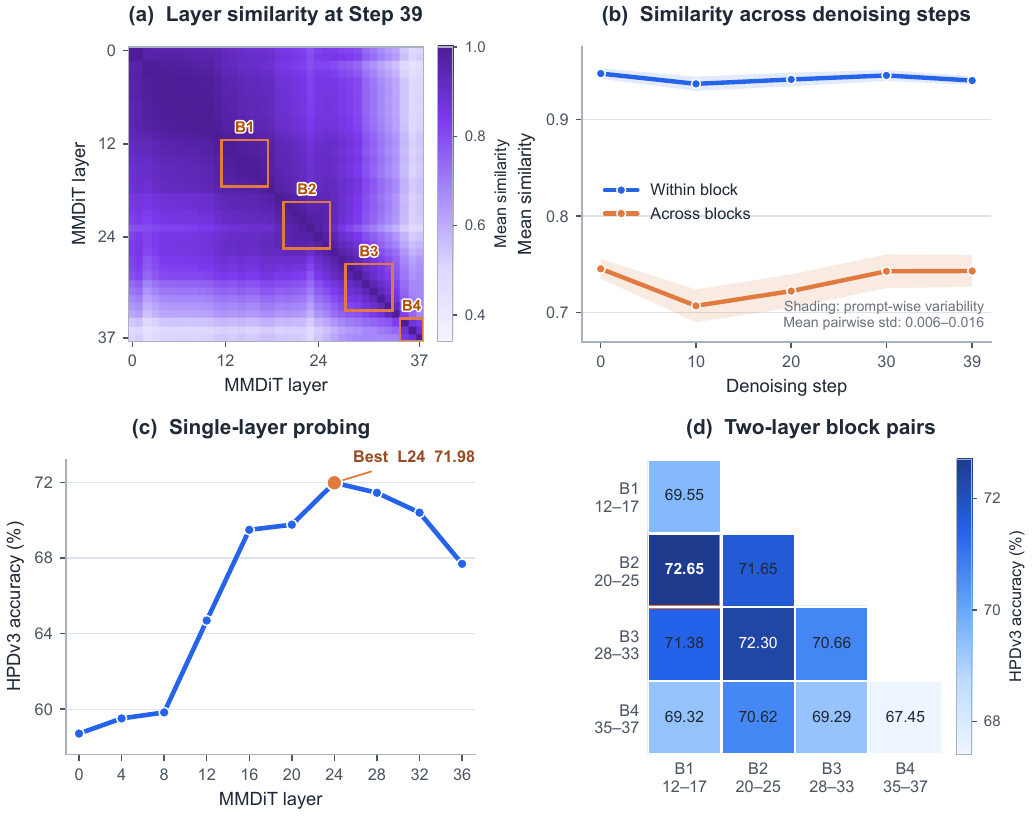}
    \caption{Layer organization and reward-relevant representations in the pretrained DiT. \textbf{(a)} Prompt-averaged layer similarity at denoising step 39, with B1--B4 denoting the layer ranges used for block-wise probing. \textbf{(b)} Mean similarity within and across blocks over denoising steps; shaded regions show prompt-wise variability. \textbf{(c)} Single-layer probing on HPDv3. \textbf{(d)} Accuracy of two-layer block pairs, with the redundant upper triangle omitted.}
    \label{fig:layer_analysis_combined}
\end{figure}

\subsection{Reward-Relevant Representations in Generative DiTs}

We first examine whether representations learned for text-to-image generation can directly support reward modeling. We freeze the pretrained SD3.5-Large DiT and train only a lightweight reward head with approximately 10M parameters on top of its intermediate image-token representations. Because these image tokens have been jointly conditioned on the prompt, this setting tests whether the frozen generative backbone already provides image--text features that are sufficiently discriminative for downstream preference prediction, rather than acquiring such features only through full reward-model fine-tuning.

\noindent\begin{minipage}{\linewidth}
\centering
\captionsetup{hypcap=false}
\captionof{table}{Effects of backbone adaptation and capacity under identical training data, reward objective, and evaluation protocol. ``Frozen'' means training only the reward head.}
\label{tab:frozen_backbone}
\small
\setlength{\tabcolsep}{3.5pt}
\begin{tabular}{lccccc}
\toprule
\textbf{Backbone} & \textbf{Backbone Training} & \textbf{ImageReward} & \textbf{PickScore} & \textbf{HPDv2} & \textbf{HPDv3} \\
\midrule
SD3.5-Medium (2.5B) & Full adaptation & 57.9          & 60.9          & 79.7          & 71.5          \\
SD3.5-Large (8.1B)  & Frozen          & 57.5          & 60.9          & 78.0          & 72.3          \\
SD3.5-Large (8.1B)  & Full adaptation & \textbf{67.0} & \textbf{66.7} & \textbf{85.6} & \textbf{77.6} \\
\bottomrule
\end{tabular}
\end{minipage}\par

As shown in Table~\ref{tab:frozen_backbone}, the head-only model reaches 57.5\%, 60.9\%, 78.0\%, and 72.3\% on ImageReward, PickScore, HPDv2, and HPDv3, respectively. A lightweight head therefore extracts substantial preference information from fixed generative representations. Full backbone adaptation adds 9.5, 5.8, 7.6, and 5.3 percentage points, showing that preference fine-tuning further strengthens the pretrained features.

We next investigate how these reward-relevant representations are organized across depth and whether different representation stages provide complementary information. Figure~\ref{fig:layer_analysis_combined} summarizes both the layer geometry and task-level probing results. Panel (a) shows the prompt-averaged layer-similarity matrix at denoising step 39. Its block structure indicates high redundancy within local layer groups and motivates the four probing regions B1--B4. Panel (b) shows that within-block similarity remains near 0.94 throughout denoising, whereas similarity across blocks remains substantially lower at 0.71--0.75. The narrow uncertainty bands are computed from prompt-wise variation. Their mean pairwise standard deviation ranges only from 0.006 to 0.016, indicating that the global layer organization is highly consistent across prompts, although variability gradually increases at later denoising steps. Complete matrices for all steps appear in Appendix~\ref{app:layer_similarity}.

To connect this representation geometry to the reward task, we train probes using individual layers and layer-block pairs. Panel (c) shows that reward accuracy is strongly depth-dependent: it rises from 58.71\% at layer 0 to 71.98\% at layer 24, before declining in the final layers. Panel (d) further shows that features from different stages are complementary. The cross-block B1+B2 pair reaches 72.65\%, outperforming the corresponding within-block B1+B1 and B2+B2 pairs. Together, these results show that a pretrained generative DiT already contains image--text representations suitable for reward modeling. The most useful information is concentrated in the middle-to-late layers while remaining distributed across distinct representation stages.

\subsection{Scaling with Generative Backbone Capacity}

Finally, we study whether DiT-based reward modeling benefits from a larger generative backbone. Under the same training data, reward objective, and evaluation protocol, increasing the backbone from SD3.5-Medium (2.5B) to SD3.5-Large (8.1B) improves ImageReward from 57.9\% to 67.0\%, PickScore from 60.9\% to 66.7\%, HPDv2 from 79.7\% to 85.6\%, and HPDv3 from 71.5\% to 77.6\%. These consistent gains indicate that larger generative backbones provide stronger representations under fixed supervision and learning objectives.

%
%
%

\section{Conclusion}
\label{sec:conclusion}
This work investigated whether text-to-image models can provide effective representations for reward modeling in addition to generating images. We introduced DiT-Reward, which converts a pretrained text-to-image DiT into a reward model. Under the same training data, DiT-Reward outperforms HPSv3 on all four preference benchmarks and achieves the best results on HPDv2 and HPDv3. Experiments that freeze the generative backbone and train only a lightweight reward head show that image representations learned by a pretrained DiT can support downstream preference prediction without further backbone adaptation. Analysis across network layers further shows that representations from the middle and later layers provide stronger reward prediction, while distinct representation stages contain complementary information. Reward performance also improves consistently as the generative backbone becomes larger. In reinforcement learning, DiT-Reward and the policy are both derived from Stable Diffusion 3.5 Large. The resulting policy consistently outperforms the HPSv3 baseline in overall score, visual quality, realism, and detail richness, while direct evaluation of generated latents enables faster reward inference with comparable memory. Although later training still exhibits weaker prompt alignment and reward hacking, our results show that pretrained generative DiTs can serve not only as image generators, but also as transferable representation backbones for evaluating generated results and optimizing related generative policies.

\bibliography{iclr2026_conference}
\bibliographystyle{iclr2026_conference}

\clearpage
\appendix
\section{DiT-Reward Training Details}
\label{sec:appendix}
\label{app:implementation_details}
\label{app:training_dynamics}

We instantiate DiT-Reward with Stable Diffusion 3.5 Large (SD3.5-L). In the official diffusers configuration, the MMDiT transformer contains 38 layers and uses 38 attention heads with head dimension 64, corresponding to a hidden dimension of 2432. The joint text-conditioning dimension is 4096, and the pooled projection dimension is 2048.

The text condition is produced by three frozen pretrained text encoders: CLIP-L/14, OpenCLIP bigG/14, and T5-XXL. The two CLIP token embeddings are concatenated along the channel dimension and padded to match the T5 embedding dimension; the resulting CLIP sequence is then concatenated with the T5 sequence along the sequence dimension and fed to the text side of the joint DiT. The pooled CLIP embeddings are also concatenated and used as global conditioning together with the timestep embedding. For image inputs, a frozen 16-channel VAE encoder maps the image into latent tokens, which are patchified and fed to the image side of the joint DiT. Under the rectified-flow formulation, the DiT takes the latent coordinate $z$, text condition $c$, and timestep $t$ as inputs and predicts a velocity field in the latent space.

\begin{figure}[htbp]
    \centering
    \includegraphics[width=\linewidth]{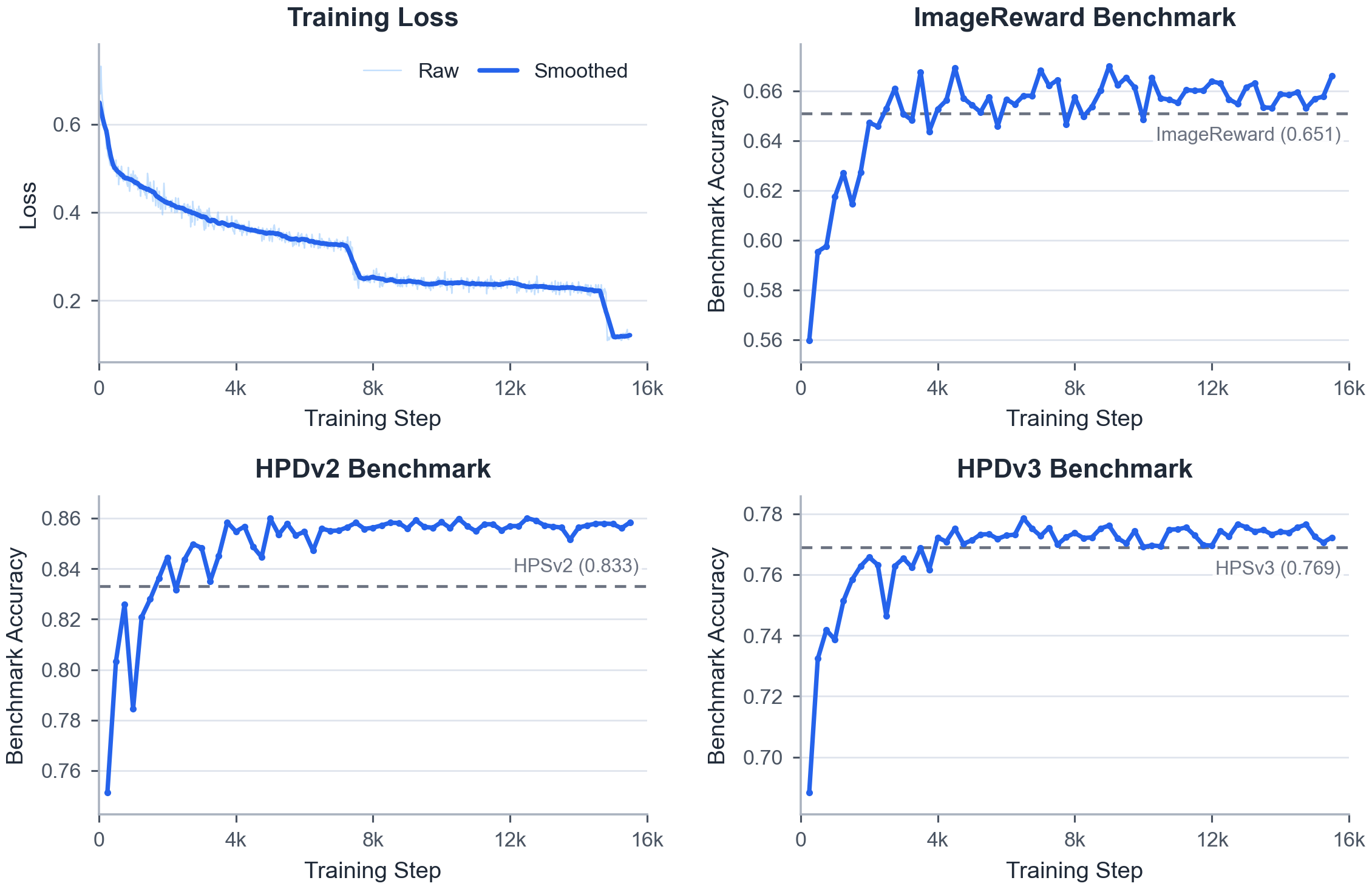}
    \caption{Training dynamics of DiT-Reward. \textbf{Top-left}: training loss. \textbf{Top-right}: ImageReward benchmark accuracy. \textbf{Bottom-left}: HPDv2 benchmark accuracy. \textbf{Bottom-right}: HPDv3 benchmark accuracy. Dashed lines indicate the corresponding baseline scores.}
    \label{fig:training_curve}
\end{figure}

\clearpage
\section{Layer Similarity Across Denoising Steps}
\label{app:layer_similarity}

Figure~\ref{fig:layer_similarity_steps_appendix} provides the complete layer-similarity statistics underlying the compact analysis in Figure~\ref{fig:layer_analysis_combined}. The prompt-averaged matrices retain a similar global block structure throughout denoising, demonstrating that the organization is not specific to a single step. Prompt-wise variability remains low relative to the similarity values, supporting the consistency of this structure across prompts. The variability nevertheless increases toward later denoising steps and is concentrated in interactions involving shallow and deep layers, suggesting increasing prompt-dependent specialization while the global layer organization remains stable.

\begin{center}
    \centering
    \includegraphics[width=\linewidth]{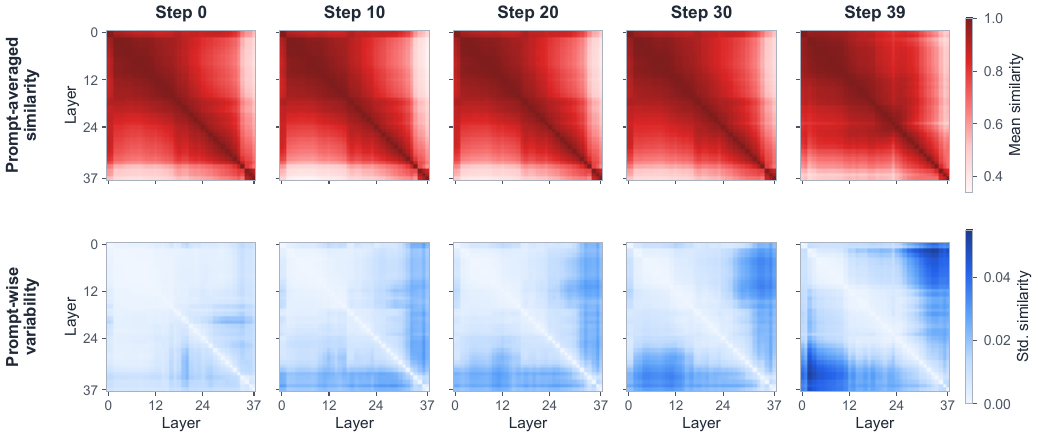}
    \captionsetup{hypcap=false}
    \captionof{figure}{Layer similarity throughout denoising. The top row shows prompt-averaged similarity among all 38 MMDiT layers, and the bottom row shows the corresponding standard deviation across prompts. Columns report denoising steps 0, 10, 20, 30, and 39.}
    \label{fig:layer_similarity_steps_appendix}
\end{center}

\clearpage
\section{Fine-Grained External VLM Evaluation During RL Training}
\label{app:rl_dimensions}

\begin{figure}[!ht]
    \centering
    \begin{subfigure}[t]{0.49\linewidth}
        \centering
        \includegraphics[width=\linewidth]{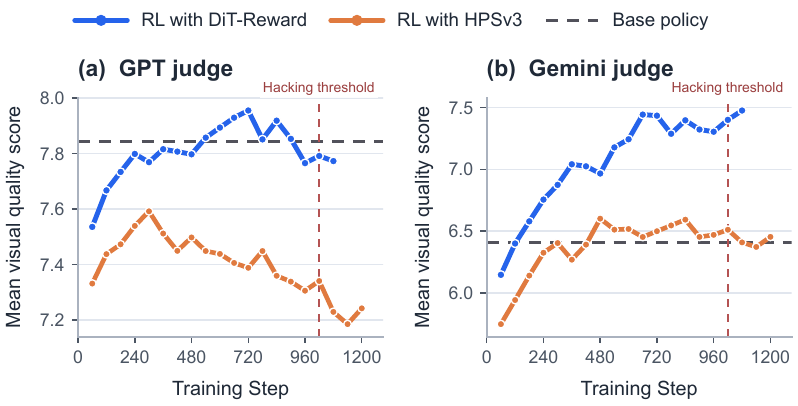}
        \caption{Visual quality}
        \label{fig:rl_visual_quality_appendix}
    \end{subfigure}
    \hfill
    \begin{subfigure}[t]{0.49\linewidth}
        \centering
        \includegraphics[width=\linewidth]{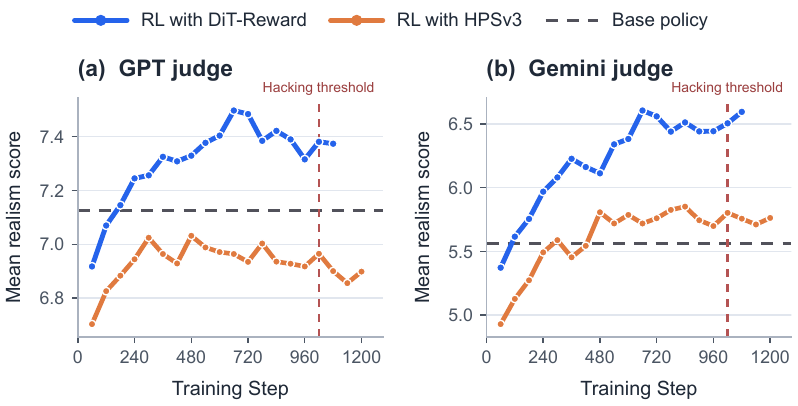}
        \caption{Realism}
        \label{fig:rl_realism_appendix}
    \end{subfigure}

    \vspace{0.5em}
    \begin{subfigure}[t]{0.49\linewidth}
        \centering
        \includegraphics[width=\linewidth]{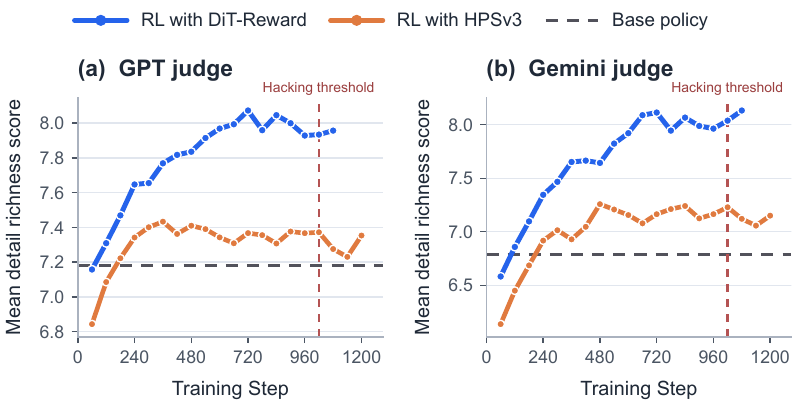}
        \caption{Detail richness}
        \label{fig:rl_detail_richness_appendix}
    \end{subfigure}
    \hfill
    \begin{subfigure}[t]{0.49\linewidth}
        \centering
        \includegraphics[width=\linewidth]{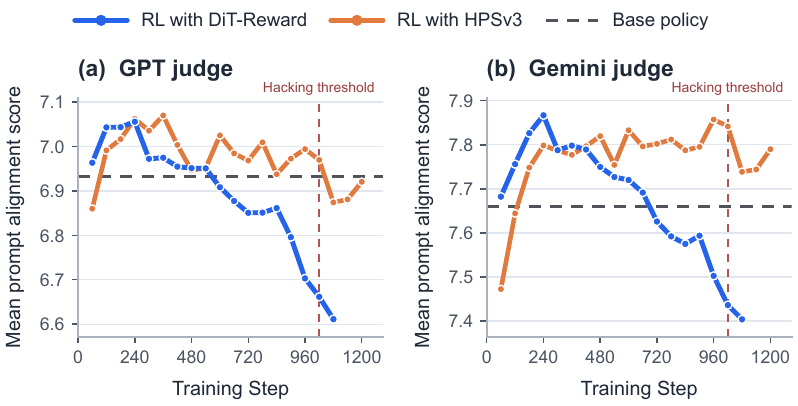}
        \caption{Prompt alignment}
        \label{fig:rl_prompt_alignment_appendix}
    \end{subfigure}
    \caption{Fine-grained external evaluation throughout RL training. Each panel reports scores from the GPT-5 and Gemini-3-Flash judges.}
    \label{fig:rl_dimensions_appendix}
\end{figure}

We use GPT-5 and Gemini-3-Flash as external judges. At each evaluation checkpoint, every generated image is evaluated independently against its text prompt. For each policy and checkpoint, we average each numeric field across the 2,048 independently evaluated images. The overall score is returned directly by the judge rather than recomputed from the four dimensions; the rationale field in the JSON response is not used in the quantitative analysis. Figure~\ref{fig:rl_dimensions_appendix} reports the resulting trajectories for visual quality, realism, detail richness, and prompt alignment.

Across the 18 matched checkpoints from steps 60 to 1080, DiT-Reward consistently outperforms the HPSv3 baseline in visual quality, realism, and detail richness. Under GPT-5, the average gains in these three dimensions are 0.39, 0.38, and 0.49 points, respectively; under Gemini-3-Flash, the corresponding gains are 0.69, 0.62, and 0.68 points. All three dimensions favor DiT-Reward at every matched checkpoint under both judges. Prompt alignment follows a different trajectory: the two policies are comparable during early training, but DiT-Reward falls behind HPSv3 at later checkpoints, yielding average differences of $-0.09$ under GPT-5 and $-0.10$ under Gemini-3-Flash. Together, these results show that the overall advantage of DiT-Reward is driven by stronger visual quality, realism, and detail richness, alongside a trade-off in prompt alignment.

Both judges receive the same system and user prompts. The system prompt is:

\begin{quote}
\footnotesize
\texttt{You are an expert evaluator for text-to-image generation models. Your task is to assess how well a generated image matches a given text prompt. You must return a JSON object and nothing else.}
\end{quote}

The user prompt is shown below, where \texttt{\{prompt\}} is replaced by the corresponding text prompt for each image:

\begin{quote}
\footnotesize
\ttfamily
Evaluate the generated image against the following text prompt:\par
\medskip
TEXT PROMPT: \char34\{prompt\}\char34\par
\medskip
Score the image on 4 dimensions (1--10, use the full range):\par
1. prompt\_alignment --- Does the image depict what the prompt describes? (subject, count, attributes, scene)\par
2. visual\_quality --- Is the image sharp, well-composed, free of artifacts?\par
3. realism --- Does it look like a plausible, coherent scene?\par
4. detail\_richness --- Are relevant details present and well-rendered?\par
\medskip
Also give an overall score (1--10, one decimal) that holistically reflects all four dimensions.\par
\medskip
Return ONLY valid JSON (no markdown, no explanation):\par
\{\par
\quad \char34 prompt\_alignment\char34: <int 1-10>,\par
\quad \char34 visual\_quality\char34: <int 1-10>,\par
\quad \char34 realism\char34: <int 1-10>,\par
\quad \char34 detail\_richness\char34: <int 1-10>,\par
\quad \char34 overall\char34: <float 1-10, one decimal>,\par
\quad \char34 rationale\char34: \char34<one sentence>\char34\par
\}
\end{quote}

\end{document}